\documentclass[sigconf]{acmart}

\usepackage{enumitem}
\setlist[itemize]{leftmargin=*}
\setlist[enumerate]{leftmargin=*}

\usepackage{todonotes}
\usepackage{booktabs}
\usepackage{multirow}
\usepackage{xcolor,colortbl}
\usepackage{xcolor}

\DeclareMathOperator{\softmax}{softmax}

\definecolor{OliveGreen}{rgb}{0,0.4,0}
\makeatletter
\newcommand*{\textoverline}[1]{$\overline{\hbox{#1}}\m@th$}
\makeatother
\definecolor{yelloww}{HTML}{149c97}
\definecolor{redd}{HTML}{59184b}

\usepackage{xspace}
\newcommand\rouge{\textsc{rouge}\xspace}
\newcommand\rougeone{\textsc{rouge-1}\xspace}

\DeclareMathOperator{\bilstm}{BiLSTM}

\usepackage{mathrsfs}

\newcommand\findings{\textsc{findings}\xspace}
\newcommand\impression{\textsc{impression}\xspace}
\newcommand\impressions{\textsc{impressions}\xspace}

\copyrightyear{2019}
\acmYear{2019}
\setcopyright{acmcopyright}
\acmConference[SIGIR '19]{Proceedings of the 42nd International ACM SIGIR Conference on Research and Development in Information Retrieval}{July 21--25, 2019}{Paris, France}
\acmPrice{15.00}
\acmDOI{10.1145/3331184.3331319}
\acmISBN{978-1-4503-6172-9/19/07}

\begin{document}

\title{Ontology-Aware Clinical Abstractive Summarization}

\author{Sean MacAvaney}
\authornote{Both authors contributed equally to this research.}
\affiliation{
 \institution{IRLab, Georgetown University}
}
\email{sean@ir.cs.georgetown.edu}
 
\author{Sajad Sotudeh}
\authornotemark[1]
\affiliation{
 \institution{IRLab, Georgetown University}
}
\email{sajad@ir.cs.georgetown.edu}
 
\author{Arman Cohan}
\affiliation{
 \institution{Allen Institute for Artificial Intelligence}
}
\email{armanc@allenai.org}

\author{Nazli Goharian}
\affiliation{
 \institution{IRLab, Georgetown University}
}
\email{nazli@ir.cs.georgetown.edu}

\author{Ish Talati}
\affiliation{
 \institution{Department of Radiology, Georgetown University}
}
\email{iat6@georgetown.edu}

\author{Ross W. Filice}
\affiliation{
 \institution{MedStar Georgetown\\ University Hospital}
}
\email{ross.w.filice@medstar.net}

\begin{abstract}
Automatically generating accurate summaries from clinical reports could save a clinician's time, improve summary coverage, and reduce errors. We propose a sequence-to-sequence abstractive summarization model augmented with domain-specific ontological information to enhance content selection and summary generation. We apply our method to a dataset of radiology reports and show that it significantly outperforms the current state-of-the-art on this task in terms of \rouge scores. Extensive human evaluation conducted by a radiologist further indicates that this approach yields summaries that are less likely to omit important details, without sacrificing readability or accuracy.
\end{abstract}

\fancyhead{}
\maketitle

\vspace{-8pt}
\section{Introduction}\label{sec:intro}

Clinical note summaries are critical to the clinical process. After writing a detailed note about a clinical encounter, practitioners often write a short summary called an \impression{} (example shown in Figure~\ref{fig:note}). This summary is important because it is often the primary document of the encounter considered when reviewing a patient's clinical history.  The summary allows for a quick view of the most important information from the report.
Automated summarization of clinical notes could save clinicians' time, and has the potential to capture important aspects of the note that the author might not have considered~\cite{gershanik2011critical}. If high-quality summaries are generated frequently, the practitioner may only need to review the summary and occasionally make minor edits.

\begin{figure}
\centering\scriptsize
\begin{tabular}{|p{7cm}|}\hline
\textbf{FINDINGS:}\\
LIVER: Liver is echogenic with slightly coarsened echotexture and mildly nodular contour. No focal lesion. Right hepatic lobe measures 14 cm in length.\\
BILE DUCTS: No biliary ductal dilatation. Common bile duct measures 0.06 cm.\\
GALLBLADDER: Partially visualized gallbladder shows multiple gallstones without pericholecystic fluid or wall thickening. ~~~~
Proximal TIPS: 108 cm/sec, previously 82 cm/sec;
Mid TIPS: 123 cm/sec, previously 118 cm/sec; 
Distal TIPS: 85 cm/sec, previously 86 cm/sec; 
PORTAL VENOUS SYSTEM:  [...]\\
\hline
\textbf{IMPRESSION: (Summary)}\\
1. Stable examination. Patent TIPS\\
2. Limited evaluation of gallbladder shows cholelithiasis.\\
3. Cirrhotic liver morphology without biliary ductal dilatation.\\
\hline
\end{tabular}
\caption{\small{Abbreviated example of radiology note and its summary.}}
\label{fig:note}
\vspace{-12pt}
\end{figure}

Recently, neural abstractive summarization models have shown successful results \cite{rush2015neural,nallapati2016abstractive,pgSee, Cohan2018ADA}.
While promising in general domains, existing abstractive models can suffer from deficiencies in content accuracy and completeness \cite{Wiseman2017ChallengesID}, which is a critical issue in the medical domain. For instance, when summarizing a clinical note, it is crucial to include all the main diagnoses in the summary accurately.
To overcome this challenge, we propose an extension to the pointer-generator model \cite{pgSee} that incorporates domain-specific knowledge for more accurate content selection. Specifically, we link entities in the clinical text with a domain-specific medical ontology (e.g., RadLex\footnote{RadLex version 3.10, \url{http://www.radlex.org/Files/radlex3.10.xlsx}} or UMLS\footnote{\url{https://www.nlm.nih.gov/research/umls/}}), and encode them into a separate context vector, which is then used to aid the generation process. We train and evaluate our proposed model on a large collection of real-world radiology \findings and \impressions from a large urban hospital, MedStar Georgetown University Hospital. Results using the \rouge evaluation metric indicate statistically significant improvements over existing state-of-the-art summarization models. Further extensive human evaluation by a radiology expert demonstrates that our method produces more complete summaries than the top-performing baseline, while not sacrificing readability or accuracy.

In summary, our contributions are:
1) An approach for incorporating domain-specific information into an abstractive summarization model, allowing for domain-informed decoding; and
2) Extensive automatic and human evaluation on a large collection of radiology notes, demonstrating the effectiveness of our model and providing insights into the qualities of our approach.

\subsection{Related Work}\label{sec:related}

Recent trends on abstractive summarization are based on sequence-to-sequence (seq2seq) neural networks with the incorporation of attention \cite{rush2015neural}, copying mechanism \cite{pgSee}, reinforcement learning objective \cite{paulus2017deep, gigioli2018domainawareabs}, and tracking coverage \cite{pgSee}. While successful, a few recent studies have shown that neural abstractive summarization models can have high readability, but fall short in generating accurate and complete content \cite{Wiseman2017ChallengesID,Gehrmann2018BottomUpAS}. Content accuracy is especially crucial in medical domain. In contrast with prior work, we focus on improving summary completeness using a medical ontology. 
\citet{gigioli2018domainawareabs} used a reinforced loss for abstractive summarization in the medical domain, although their focus was headline generation from medical literature abstracts. Here, we focus on summarization of clinical notes where content accuracy and completeness are more critical. The most relevant work to ours is by \citet{Zhang2018LearningTS} where an additional section from the radiology report (\textsc{background}) is used to improve summarization. 
Extensive automated and human evaluation and analyses demonstrate the benefits of our proposed model in comparison with existing work.

\section{Model}\label{sec:method}

\textbf{Pointer-generator network (PG).}
Standard neural approaches for abstractive summarization follow the seq2seq framework where an encoder network reads the input and a separate decoder network (often augmented with an attention mechanism) learns to generate the summary ~\cite{sutskever2014sequence}. Bidirectional LSTMs (BiLSTMs) \cite{hochreiter1997long} are often used as the encoder and decoder.
A more recent successful summarization model---called Pointer-generator network---allows the decoder to also directly copy text from the input in addition to generation ~\cite{pgSee}. Given a report $\mathbf{x}=\{x_1,x_2,...,x_n\}$, the encoded input sequence $\mathbf{h}=\bilstm(\mathbf{x})$, and the current decoding state $\mathbf{s_t}=\bilstm(\mathbf{x}^\prime)[t]$, where $\mathbf{x}^\prime$ is the input to the decoder (i.e., gold standard summary token at training or previously generated token at inference time), the model computes the attention weights over the input terms $\mathbf{a}=\softmax(\mathbf{\mathbf{h^\top \mathbf{W_1} s^\top}})$. The attention scores are employed to compute a context vector $c$ which is a weighted sum over input $\mathbf{c}=\sum_i^n a_i\mathbf{h}_i$ that is used along with the output of the decoder $\bilstm$ to either generate the next term from a known vocabulary or copy the token from the input sequence with the highest attention value. We refer the reader to \citet{pgSee} for additional details on the pointer-generator architecture.

\textbf{Ontology-aware pointer-generator (Ontology PG).}
In this work, we propose an extension of the pointer-generator network that allows us to leverage domain-specific knowledge encoded in an ontology to improve clinical summarization. 
We introduce a new encoded sequence $\mathbf{u}=\{u_1,...,u_{n^\prime}\}$ that is the result of linking an ontology $\mathscr{U}$ to the input texts. In other words, $\mathbf{u}=F_\mathscr{U}(\mathbf{x})$ where $F_\mathscr{U}$ is a mapping function, e.g., a simple mapping function that only outputs a word sequence if it appears in the ontology and otherwise skips it. We then use a second $\bilstm$ to encode this additional ontology terms similar to the way the original input is encoded $\mathbf{h}_u=BiLSTM(\mathbf{u})$.
We then calculate an additional context vector $\mathbf{c}^\prime$ which includes the domain-ontology information:
\begin{equation}
    \mathbf{a}^\prime=\softmax(\mathbf{\mathbf{h_u^\top \mathbf{W}_2 \mathbf{s}^\top}}); \;\;\; \mathbf{c}^\prime = \sum\nolimits_i^{n^\prime} a^\prime_i \mathbf{u}_i
\end{equation}
The second context vector acts as additional global information to aid the decoding process, and is akin to how~\citet{Zhang2018LearningTS} include \textsc{background} information from the report. We modify the decoder BiLSTM to include the ontology-aware context vector in the decoding process. Recall that an LSTM network controls the flow of its previous state and the current input using several gates (input gate \textbf{i}, forget gate \textbf{f}, and output gate \textbf{o}), where each of these gates are vectors calculated according to an additive combination of the previous LSTM state and current input. For example, for the forget gate we have: $\mathbf{f}_t=\tanh(W_{f}[s_{t-1}; x^\prime_t]+b)$ 
where $s_{t-1}$ is the previous decoder state and $x^\prime_t$ is the decoder input, and ``;'' shows concatenation (for more details on LSTMs refer to \cite{hochreiter1997long}). The ontology-aware context vector $c^\prime$ is passed as additional input to this function for all the LSTM gates: e.g., for the forget gate we will have: $\mathbf{f}_t=\tanh(W_{f}[s_{t-1}; x^\prime_t; c^\prime]+b)$. This intuitively guides the information flow in the decoder using the ontology information.

\section{Experimental setup}\label{sec:exp-setup}
We train and evaluate our model on a dataset of 41,066 real-world radiology reports from MedStar Georgetown University Hospital containing radiology reports with a variety of imaging modalities (e.g., x-rays, CT scans, etc). The dataset is randomly split into 80-10-10 train-dev-test splits. Each report describes clinical \findings{} about a specific diagnostic case, and an \impression{} summary (as shown in Figure~\ref{fig:note}). The \findings{} sections are 136.6 tokens on average and the \impression{} sections are 37.1 tokens on average.
Performing cross-institutional evaluation is challenging and beyond the scope of this work due to the varying nature of reports between institutions. For instance, the public Indiana University radiology dataset~\cite{demner2015preparing} consists only of chest x-rays, and has much shorter reports (average length of \findings{}: 40.0 tokens; average length of \impressions{}: 10.5 tokens). Thus, in this work, we focus on summarization within a single institution.

\textbf{Ontologies.}
We employ two ontologies in this work. UMLS is a general medical ontology maintained by the US National Library of Medicine and includes various procedures, conditions, symptoms, body parts, etc. We use QuickUMLS~\cite{soldaini2016quickumls} (a fuzzy UMLS concept matcher) with a Jaccard similarity threshold of 0.7 and a window size of 3 to extract UMLS concepts from the radiology \findings.
We also evaluate using an ontology focused on radiology, RadLex, which is a widely-used ontology of radiological terms maintained by the Radiological Society of North America. It consists of 68,534 radiological concepts organized according to a hierarchical structure. We use exact n-gram matching to find important radiological entities, only considering RadLex concepts at a depth of 8 or greater.\footnote{The maximum tree depth is 20.} In pilot studies, we found that the entities between depths 8 and 20 tend to represent concrete entities (e.g., `thoracolumbar spine region') rather than abstract categories (e.g., `anatomical entity').

\begin{figure*}
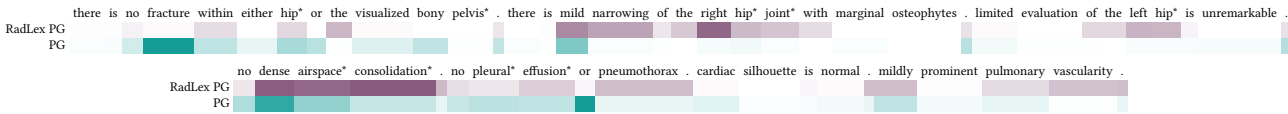

\tiny
\bgroup
\def\arraystretch{1}
\setlength{\tabcolsep}{0.3em}

\begin{tabular}{rlllllllllllllllllllllllllllllllllll}
&there&is&no&fracture&within&either&hip*&or&the&visualized&bony&pelvis*&.&there&is&mild&narrowing&of&the&right&hip*&joint*&with&marginal&osteophytes&.&limited&evaluation&of&the&left&hip*&is&unremarkable&.\\
RadLex PG&\cellcolor{redd!1}&\cellcolor{redd!1}&\cellcolor{redd!6}&\cellcolor{redd!2}&\cellcolor{redd!15}&\cellcolor{redd!0}&\cellcolor{redd!17}&\cellcolor{redd!0}&\cellcolor{redd!31}&\cellcolor{redd!2}&\cellcolor{redd!0}&\cellcolor{redd!0}&\cellcolor{redd!11}&\cellcolor{redd!0}&\cellcolor{redd!1}&\cellcolor{redd!51}&\cellcolor{redd!40}&\cellcolor{redd!11}&\cellcolor{redd!24}&\cellcolor{redd!66}&\cellcolor{redd!30}&\cellcolor{redd!26}&\cellcolor{redd!15}&\cellcolor{redd!0}&\cellcolor{redd!0}&\cellcolor{redd!13}&\cellcolor{redd!2}&\cellcolor{redd!0}&\cellcolor{redd!17}&\cellcolor{redd!17}&\cellcolor{redd!33}&\cellcolor{redd!32}&\cellcolor{redd!5}&\cellcolor{redd!0}&\cellcolor{redd!13}\\
PG&\cellcolor{yelloww!2}&\cellcolor{yelloww!3}&\cellcolor{yelloww!19}&\cellcolor{yelloww!100}&\cellcolor{yelloww!28}&\cellcolor{yelloww!10}&\cellcolor{yelloww!37}&\cellcolor{yelloww!30}&\cellcolor{yelloww!1}&\cellcolor{yelloww!14}&\cellcolor{yelloww!27}&\cellcolor{yelloww!1}&\cellcolor{yelloww!26}&\cellcolor{yelloww!2}&\cellcolor{yelloww!1}&\cellcolor{yelloww!55}&\cellcolor{yelloww!2}&\cellcolor{yelloww!1}&\cellcolor{yelloww!0}&\cellcolor{yelloww!4}&\cellcolor{yelloww!6}&\cellcolor{yelloww!2}&\cellcolor{yelloww!0}&\cellcolor{yelloww!2}&\cellcolor{yelloww!0}&\cellcolor{yelloww!30}&\cellcolor{yelloww!5}&\cellcolor{yelloww!1}&\cellcolor{yelloww!1}&\cellcolor{yelloww!0}&\cellcolor{yelloww!3}&\cellcolor{yelloww!3}&\cellcolor{yelloww!3}&\cellcolor{yelloww!4}&\cellcolor{yelloww!32}\\
\end{tabular}

\vspace{0.6em}

\begin{tabular}{rlllllllllllllllllllll}
&no&dense&airspace*&consolidation*&.&no&pleural*&effusion*&or&pneumothorax&.&cardiac&silhouette&is&normal&.&mildly&prominent&pulmonary&vascularity&.\\
RadLex PG&\cellcolor{redd!11}&\cellcolor{redd!70}&\cellcolor{redd!65}&\cellcolor{redd!71}&\cellcolor{redd!30}&\cellcolor{redd!14}&\cellcolor{redd!11}&\cellcolor{redd!21}&\cellcolor{redd!4}&\cellcolor{redd!28}&\cellcolor{redd!28}&\cellcolor{redd!2}&\cellcolor{redd!0}&\cellcolor{redd!4}&\cellcolor{redd!2}&\cellcolor{redd!28}&\cellcolor{redd!28}&\cellcolor{redd!1}&\cellcolor{redd!16}&\cellcolor{redd!27}&\cellcolor{redd!28}\\
PG&\cellcolor{yelloww!34}&\cellcolor{yelloww!88}&\cellcolor{yelloww!46}&\cellcolor{yelloww!25}&\cellcolor{yelloww!10}&\cellcolor{yelloww!24}&\cellcolor{yelloww!29}&\cellcolor{yelloww!27}&\cellcolor{yelloww!100}&\cellcolor{yelloww!10}&\cellcolor{yelloww!9}&\cellcolor{yelloww!13}&\cellcolor{yelloww!2}&\cellcolor{yelloww!3}&\cellcolor{yelloww!5}&\cellcolor{yelloww!9}&\cellcolor{yelloww!27}&\cellcolor{yelloww!5}&\cellcolor{yelloww!10}&\cellcolor{yelloww!1}&\cellcolor{yelloww!9}\\
\end{tabular}

\vspace{0.6em}

\caption{{Average attention weight comparison between our approach (RadLex PG) and the baseline (PG). Color differences show to which term each model attends more while generating summary. RadLex concepts of depth 8 or lower are marked with~*. Our approach attends to more RadLex terms throughout the document, allowing for more complete summaries.}}
\label{fig:attn}
\egroup
\end{figure*}

\textbf{Comparison.}
We compare our model to well-established extractive baselines as well as the state-of-the-art abstractive summarization models.
\vspace{-0.38em}
\begin{itemize}[leftmargin=*,label={-}]
\item\textbf{LSA}~\cite{Steinberger2004LSA}: An extractive vector-space summarization model based on Singular Value Decomposition (SVD). 
\item \textbf{LexRank}~\cite{erkanLexrank}: An extractive method which employs graph-based centrality ranking of the sentence.\footnote{For LSA and LexRank, we use the Sumy implementation (\url{https://pypi.python.org/pypi/sumy}) with the top 3 sentences.} 
\item \textbf{Pointer-Generator (PG)}~\cite{pgSee}: An abstractive seq2seq attention summarization model that incorporates a copy mechanism to directly copy text from input where appropriate.

\item \textbf{Background-Aware Pointer-Generator (Back. PG)} ~\cite{Zhang2018LearningTS}: An extension of PG, which is specifically designed to improve radiology note summarization by encoding the \textsc{Background} section of the report to aid the decoding process.\footnote{Using the author's code at \url{github.com/yuhaozhang/summarize-radiology-findings}}
\end{itemize}

\textbf{Parameters and training.}\label{sec:exp-setup}
We use 100-dimensional GloVe embeddings pre-trained over a large corpus of 4.5 million radiology reports~\cite{Zhang2018LearningTS}, a 2-layer $\bilstm$ encoder with a hidden size of 100, and a 1-layer LSTM decoder with the hidden size of 200. At inference time, we use beam search with beam size of 5. We use a dropout of 0.5 in all models, and train to optimize negative log-likelihood loss using the Adam optimizer~\cite{Diedrik2014Adam} and a learning rate of 0.001. 

\begin{table}
\centering\small
\caption{\rouge results on MedStar Georgetown University Hospital's development and test sets. Both the UMLS and RadLex ontology PG models are statistically better than the other models (paired t-test, $p<0.05$).}
\begin{tabular}{lrrrrrr}
\toprule
&\multicolumn{3}{c}{Development}&\multicolumn{3}{c}{Test}\\
\cmidrule(lr){2-4}\cmidrule(lr){5-7}
Model & RG-1 & RG-2 & RG-L & RG-1 & RG-2 & RG-L \\
\midrule
LexRank~\cite{erkanLexrank} & 27.60 & 13.85 & 25.79 & 28.02 & 14.26 & 26.24 \\
LSA~\cite{Steinberger2004LSA} & 28.04 & 14.68 & 26.15 & 28.16 & 14.71 & 26.27 \\
PG~\cite{pgSee} & 36.60 & 21.73 & 35.40 & 37.17 & 22.36 & 35.45 \\
Back. PG~\cite{Zhang2018LearningTS} & 36.58 & 21.86 & 35.39 & 36.95 & 22.37 & 35.68 \\
\midrule
UMLS PG (ours) & 37.41 & 22.23 & 36.10 & 37.98 & 23.14 &  36.67 \\
RadLex PG (ours) & \textbf{37.64} & \textbf{22.45} & \textbf{36.33} &  \textbf{38.42} & \textbf{23.29} & \textbf{37.02} \\
\bottomrule
\end{tabular}
\label{tab:results_medstar}
\end{table}

\section{Results and Analysis}\label{sec:results}
\subsection{Experimental results}
Table~\ref{tab:results_medstar} presents \rouge evaluation results of our model compared with the baselines (as compared to human-written \impressions). The extractive summarization methods (LexRank and LSA) perform particularly poorly. This may be due to the fact that these approaches are limited to simply selecting sentences from the text, and that the most central sentences may not be the most important for building an effective \impression{} summary.
Interestingly, the Back. PG approach (which uses the \textsc{background} section of the report to guide the decoding process) is ineffective on our dataset. This may be due to differences in conventions across institutions, such as what information is included in a report's \textsc{background} and what is considered important to include in its \impression.

We observe that our Ontology-Aware models (UMLS PG and RadLex PG) significantly outperform all other approaches (paired t-test, $p<0.05$) on both the development and test sets. The RadLex model slightly outperforms the UMLS model, suggesting that the radiology-specific ontology is beneficial (though the difference between UMLS and RadLex is not statistically significant). We also experimented incorporating both ontologies in the model simultaneously, but it resulted in slightly lower performance (1.26\% lower than the best model on \rougeone). To verify that including ontological concepts in the decoder helps the model identify and focus on more radiology terms, we examined the attention weights. In Figure~\ref{fig:attn}, we show attention plots for two reports, comparing the attention of our approach and PG. The plots show that our approach results in attention weights being shared across radiological terms throughout the \findings{}, potentially helping the model to capture a more complete summary.

\begin{figure*}
\centering
\bgroup
\setlength{\tabcolsep}{0pt}
\renewcommand{\arraystretch}{0}
\begin{tabular}{ccc|ccc}
\includegraphics[scale=0.27]{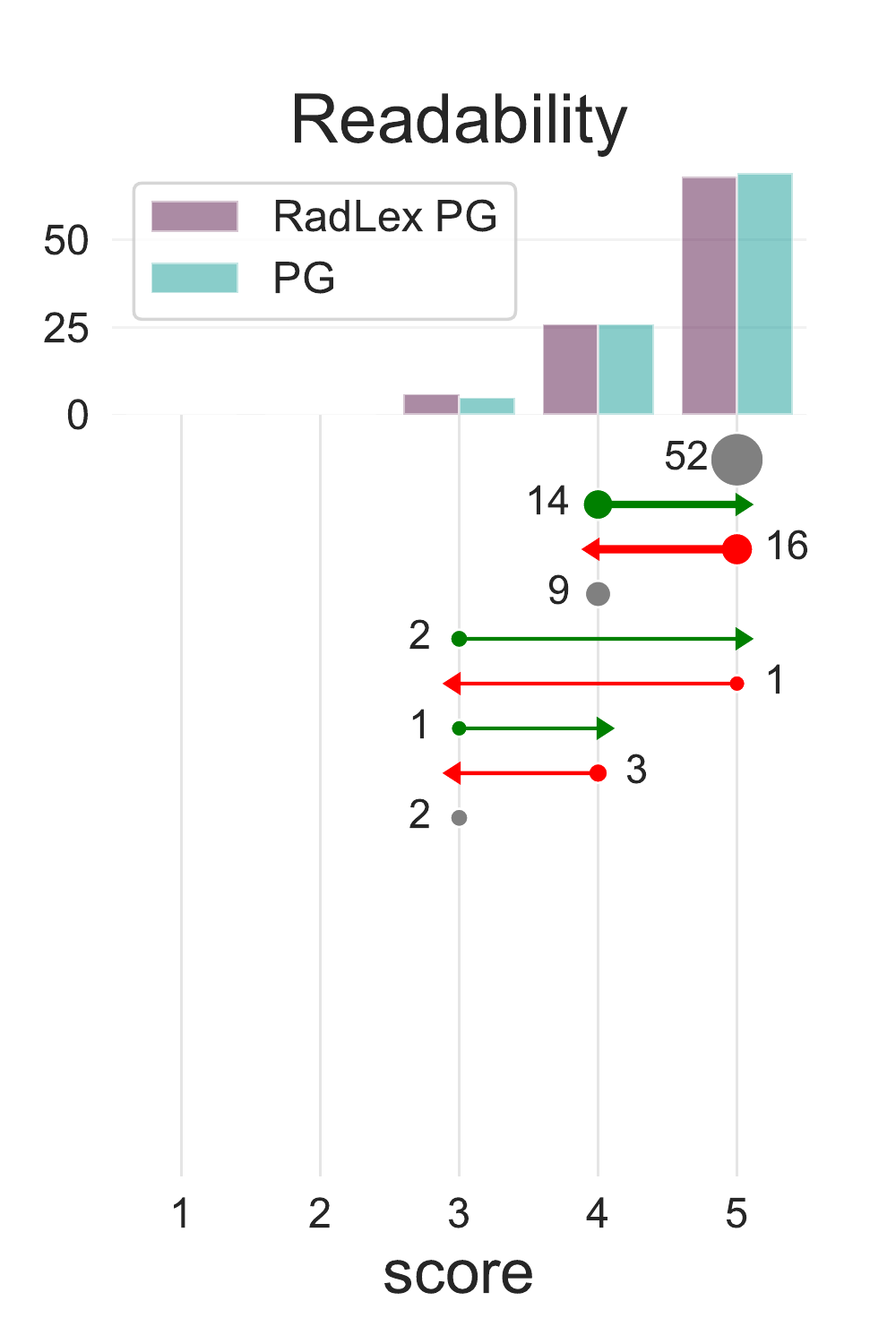} &
\includegraphics[scale=0.27]{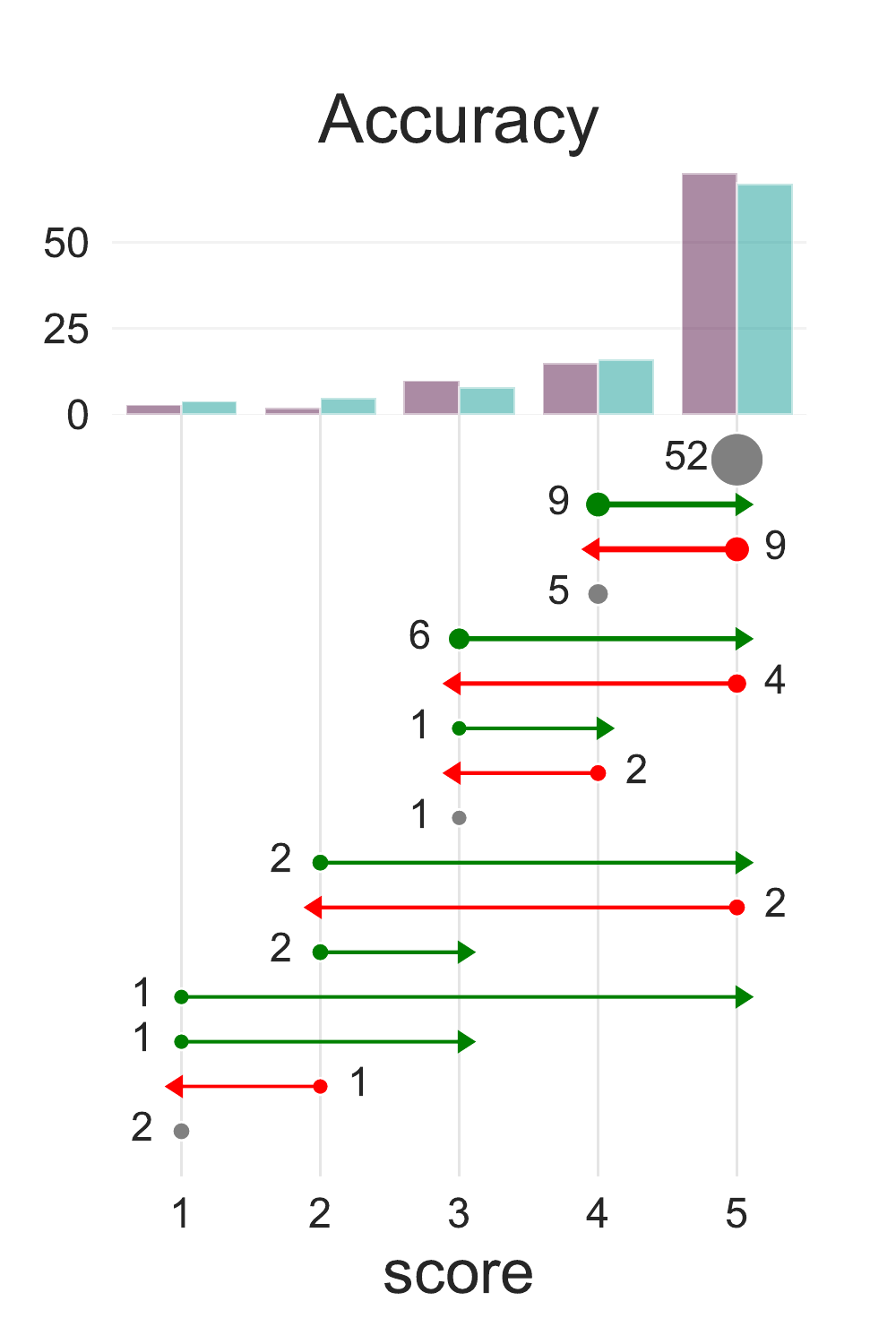} &
\includegraphics[scale=0.27]{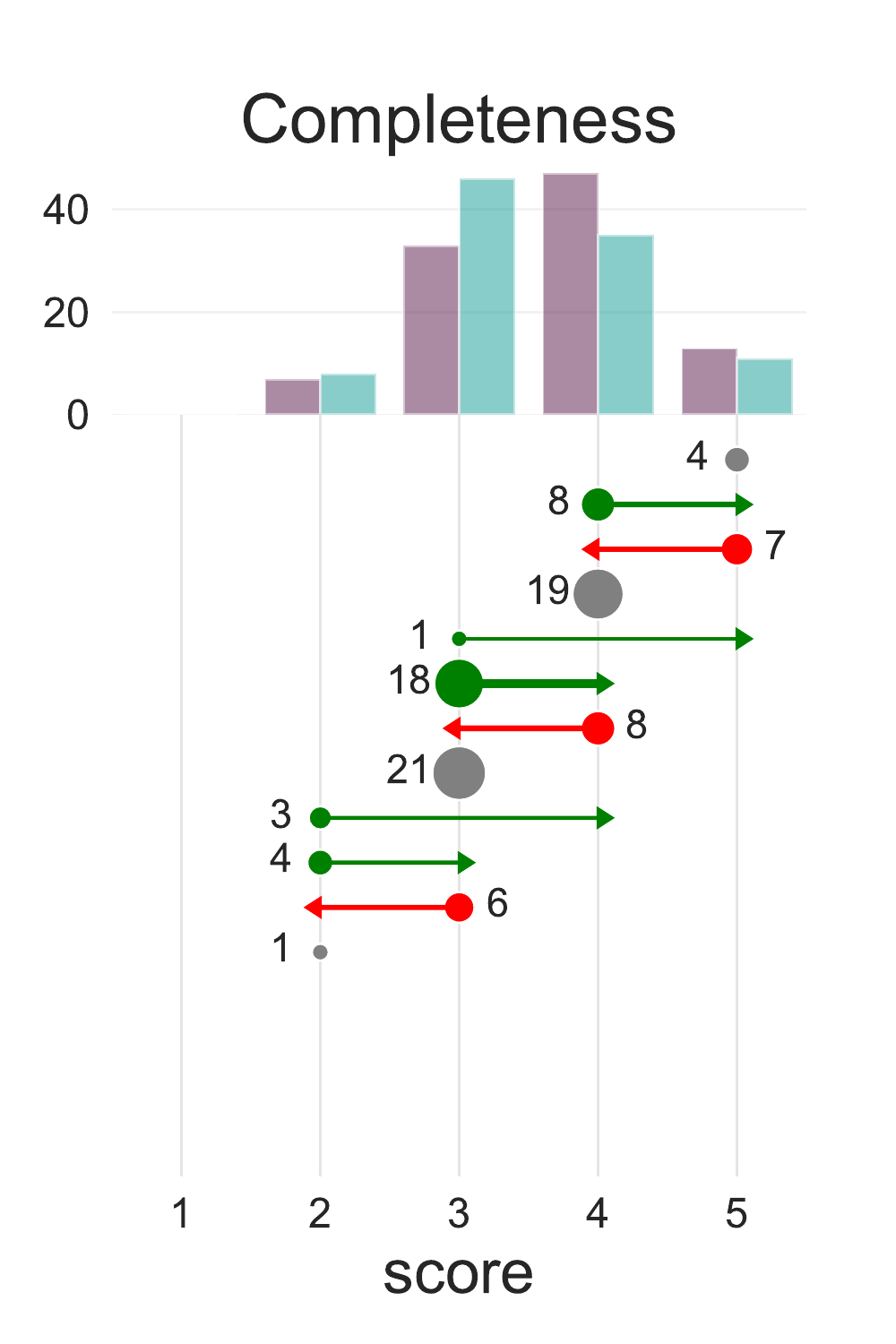} &
\includegraphics[scale=0.27]{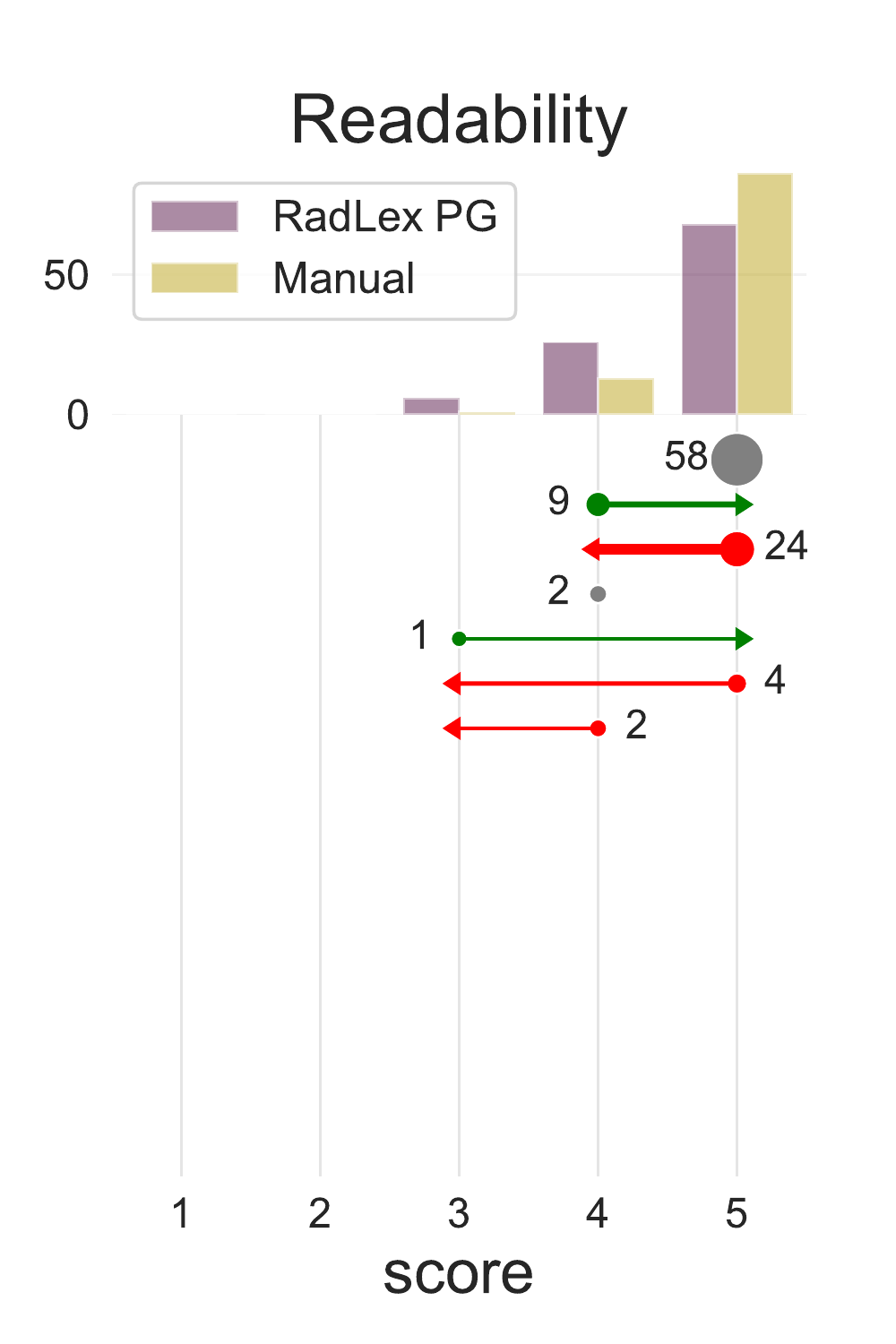} &
\includegraphics[scale=0.27]{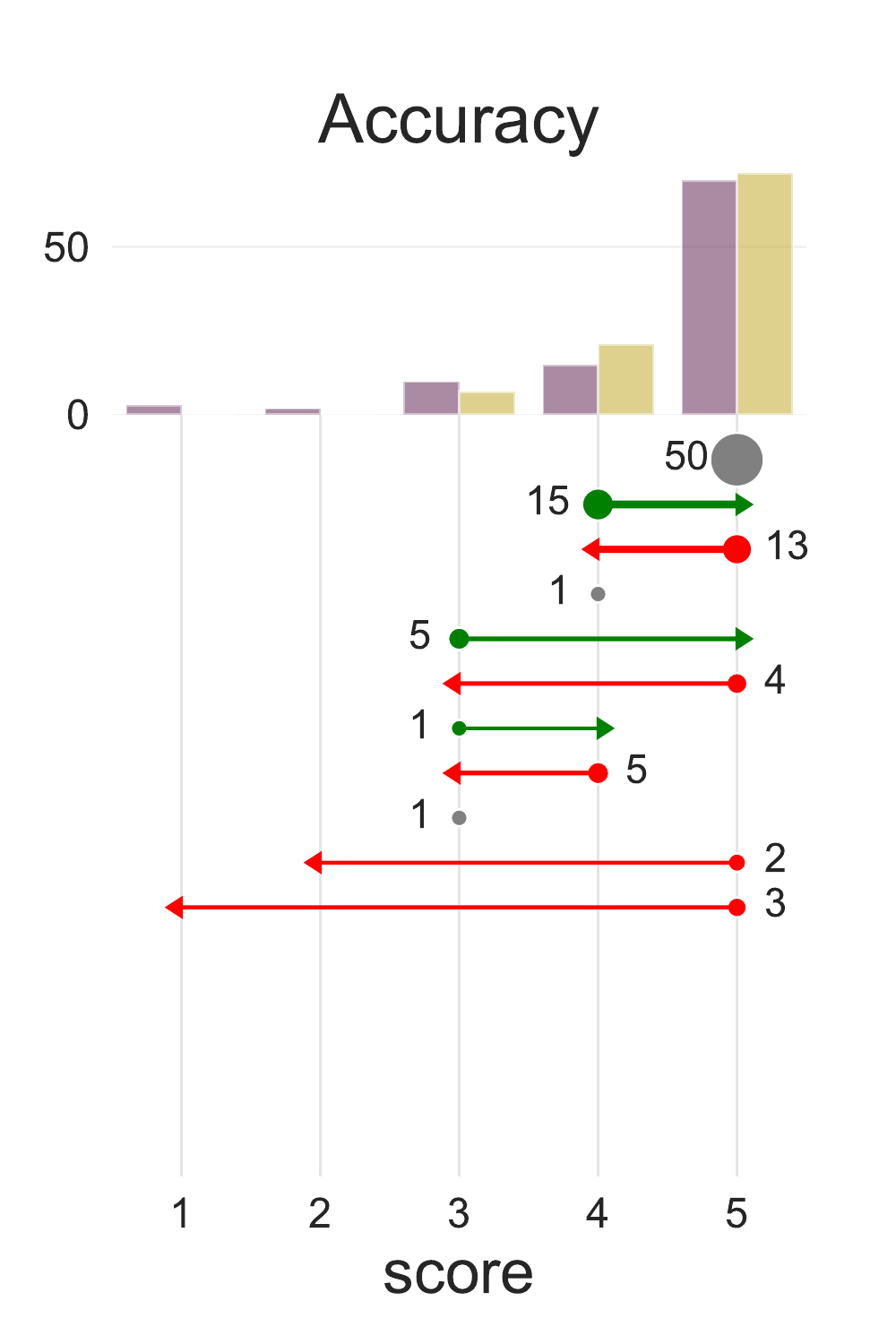} &
\includegraphics[scale=0.27]{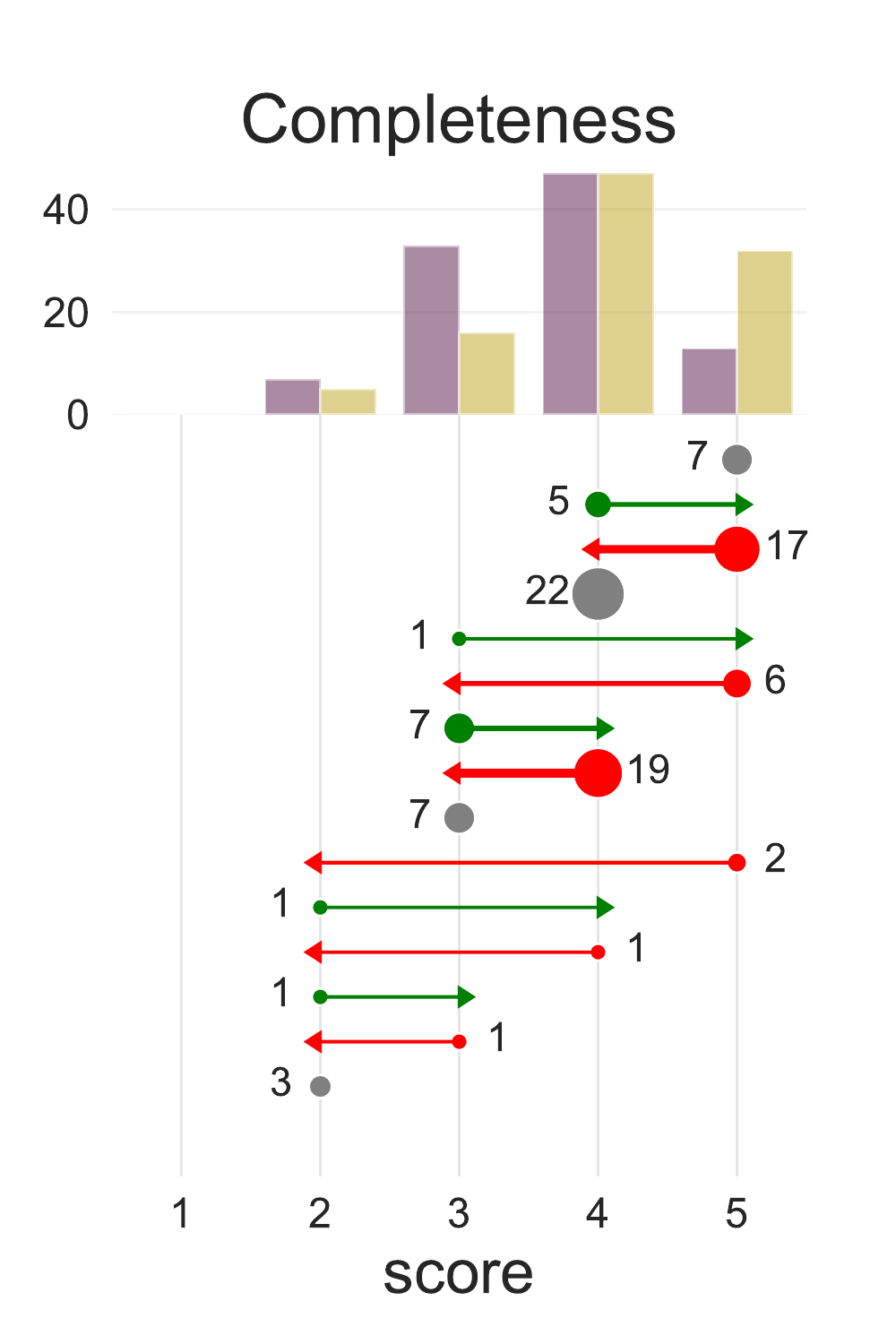} \\
(a) & (b) & (c) & (d) & (e) & (f) \\
\end{tabular}
\vspace{-0.5em}
\caption{Histograms and arrow plots plot depicting differences between \impressions{} of 100 manually-scored radiology reports. Although challenges remain to reach human parity for all metrics, our approach makes strong gains to address the problem of report completeness (c,~f), as compared to the next leading summarization approach (PG).}
\label{fig:arrows}
\egroup
\end{figure*}

\subsection{Expert human evaluation}

While our approach surpasses state-of-the-art results on our dataset in terms of \rouge scores, we recognize the limitations of the \rouge framework for evaluating summarization \cite{Conroy2008MindTG,Cohan2016RevisitingSE}. To gain better insights into how and why our methodology performs better, we also conduct expert human evaluation. We had a domain expert (radiologist) who is familiar with the process of writing radiological \findings{} and \impressions{} evaluate 100 reports. Each report consists of the radiology \findings, one manually-written \impression, one \impression generated using PG, and one \impression generated using our ontology PG method (with RadLex). In each sample, the order of the \textsc{Impression}s  are shuffled to avoid bias between samples. Samples were randomly chosen from the test set, one from each of 100 evenly-spaced bins sorted by our system's \rougeone score.
The radiologist was asked to score each \impression{} in terms of the following on a scale of 1 (worst) to 5 (best):
\vspace{-0.3em}
\begin{enumerate}[leftmargin=3mm,label={-}]
\item \textbf{Readability.} Impression is understandable (5) or gibberish (1).
\item \textbf{Accuracy.} Impression is fully accurate (5), or contains critical errors (1).
\item \textbf{Completeness.} Impression contains all important information (5), or is missing important points (1).
\end{enumerate}
\vspace{-0.3em}

We present our manual evaluation results using histograms and arrow plots in Figure~\ref{fig:arrows}. The histograms indicate the score distributions of each approach, and the arrows indicate how the scores changed. The starting points of an arrow indicates the score of an \impression{} we compare to (either the human-written, or the summary generated by PG). The head of an arrow indicates the score of our approach. The numbers next to each arrow indicate how many reports made the transition. The figure shows that our approach improves completeness considerably, while maintaining the readability and accuracy. The major improvement in completeness is between the score of 3 and 4, where there is a net gain of 10 reports. Completeness is particularly important because it is where existing summarization models---such as PG---are currently lacking, as compared to human performance. Despite the remaining gap between human and generated completeness, our approach yields considerable gains toward human-level completeness. Our model is nearly as accurate as human-written summaries, only making critical errors (scores of 1 or 2) in 5\% of the cases evaluated, as compared to 8\% of cases for PG. No critical errors were found in the human-written summaries, although the human-written summaries go through a manual review process to ensure accuracy.

The expert annotator furthermore conducted blind qualitative analysis to gain a better understanding of when our model is doing better and how it can be further improved. In line with the completeness score improvements, the annotator noted that in many cases our approach is able to identify pertinent points associated with RadLex terms that were missed by the PG model. In some cases, such as when the author picked only one main point, our approach was able to pick up important items that the author missed. Interestingly, it also was able to include specific measurement details better than the PG network, even though these measurements do not appear in RadLex. Although readability is generally strong, our approach sometimes generates repetitive sentences and syntactical errors more often than humans. These could be addressed in future work with additional post-processing heuristics such as removing repetitive n-grams as done in \cite{paulus2017deep}. In terms of accuracy, our approach sometimes mixes up the ``left'' and ``right'' sides. This often occurs with \findings that have mentions of both sides of a specific body part.
Multi-level attention (e.g., \cite{Cohan2018ADA}) could address this by forcing the model to focus on important segments of the text.
There were also some cases where our model under-performed in terms of accuracy and completeness due to synonymy that is not captured by RadLex. For instance, in one case our model did identify torsion, likely due to the fact that in the \findings{} section it was referred to as \textit{twisting} (a term that does not appear in RadLex).

\section{Conclusion}\label{sec:conclusion}

In this work, we present an approach for informing clinical summarization models of ontological information. This is accomplished by providing an encoding of ontological terms matched in the original text as an additional feature to guide the decoding. We find that our system exceeds state-of-the-art performance at this task, producing summaries that are more comprehensive than those generated by other methods, while not sacrificing readability or accuracy.

\section*{Acknowledgments}
We thank the additional residents, Lee McDaniel and Marlie Philiossaint, for their contributive evaluations, as well as anonymous reviewers for their useful feedback.

\bibliographystyle{ACM-Reference-Format}
\bibliography{sample-base}

\end{document}